\documentclass[10pt]{article} 
\usepackage[preprint]{tmlr}

\usepackage{amsmath,amsfonts,bm}

\def\eqref#1{equation~\ref{#1}}

\def\1{\bm{1}}

\DeclareMathAlphabet{\mathsfit}{\encodingdefault}{\sfdefault}{m}{sl}
\SetMathAlphabet{\mathsfit}{bold}{\encodingdefault}{\sfdefault}{bx}{n}

\usepackage{hyperref}
\usepackage{url}
\usepackage{nicematrix}
\usepackage{enumitem}

\title{The Effectiveness of Approximate Regularized Replay for Efficient Supervised Fine-Tuning of Large Language Models}

\author{\name Matthew Riemer$^{1,2}$, Erik Miehling$^1$, Miao Liu$^1$, Djallel Bouneffouf$^1$, Murray Campbell$^1$  \\
      \addr $^1$IBM Research $^2$Mila, Universit\'e de Montr\'eal \\
      Contact: mdriemer@us.ibm.com}

\def\month{MM}  
\def\year{YYYY} 
\def\openreview{\url{https://openreview.net/forum?id=XXXX}} 

\begin{document}

\maketitle

\begin{abstract}
Although parameter-efficient fine-tuning methods, such as LoRA, only modify a small subset of parameters, they can have a significant impact on the model. Our instruction-tuning experiments show that LoRA-based supervised fine-tuning can catastrophically degrade model capabilities, even when trained on very small datasets for relatively few steps. With that said, we demonstrate that while the most straightforward approach (that is likely the most used in practice) fails spectacularly, small tweaks to the training procedure with very little overhead can virtually eliminate the problem. Particularly, in this paper we consider a regularized approximate replay approach which penalizes KL divergence with respect to the initial model and interleaves in data for next token prediction from a different, yet similar, open access corpus to what was used in pre-training. When applied to Qwen instruction-tuned models, we find that this recipe preserves general knowledge in the model without hindering plasticity to new tasks by adding a modest amount of computational overhead. 
\end{abstract}

\section{Introduction}

The problem of continual learning with neural networks has remained a challenging problem for over three decades \citep{ring1994continual}. Indeed, it is well known that continual learners face a dilemma between prioritizing stability and plasticity in the weights of the model \citep{StabilityPlasticity}. It is also well documented that when neural networks perform optimization for extended periods on only a single task, they tend to experience the phenomena of \textit{catastrophic forgetting} \citep{CF} in which plasticity of the network to the current task leads to substantial reduction in the quality of the model on prior tasks encountered during training. While many different settings have been identified in the literature under the umbrella of continual learning \citep{normandin2021sequoia}, simple instruction-tuning with LLMs is not often considered one of them. As a result, we believe that many practitioners have been delving into the waters of continual learning without even realizing it. The truth is that simple supervised fine-tuning with LLMs, even on small datasets with relatively few steps of optimization, is a setting under the umbrella of continual learning for which we have a strong expectation that the model will experience catastrophic forgetting of past capabilities. We believe that LLMs are rarely evaluated by practitioners for their general capabilities after instruction-tuning, and that this could be a contributing reason for the high failure rate of recent projects to get return for their business from generative AI \citep{Challapally2025GenAIDivide}. 

Recent studies have demonstrated that catastrophic forgetting is indeed a significant problem during fine-tuning of LLMs \citep{luo2025empirical,Li2024RevisitingCatastrophicForgetting,Zhai2024InvestigatingCatastrophicForgettingMLLMs}. However, these studies focused on full fine-tuning rather than parameter efficient fine-tuning methods such as LoRA \citep{hu2022lora} that have become quite popular because of their increased computational and memory efficiency. Intuitively, because LoRA adapters only learn very few parameters relative to the base model, it is often believed that forgetting is less of an issue with these models. Our experiments directly contradict this narrative and demonstrate that forgetting is still a very substantial issue during training with LoRA -- even when the trainable parameters are less than 1\% of the size of the base model. Recent papers have considered modifications to LoRA's decomposition itself in order to prevent forgetting \citep{Fawi2024CURLoRA,Xiong2025OPLoRA,Lu2025CLoRA}. In contrast, we consider simple methods to address forgetting that are agnostic to the particular fine-tuning method leveraged, which we show to also be effective when training with LoRA. 

The recent work of \citet{shenfeld2025rl} demonstrated that RL training of LLMs results is considerably less forgetting than is experienced during supervised fine-tuning. There are two reasons provided by the authors for this insight: 1) RL when applied to LLMs is generally KL regularized to prevent drift from the initial model parameters, and 2) RL training leverages on-policy data to update the model. While we definitely agree with \citet{shenfeld2025rl}'s insight that these two things are very synergistic when applied together, we also feel that it is worth considering how much KL regularization can improve supervised fine-tuning on its own. While it is not often applied in supervised fine-tuning settings, we find that there are hyperparameter values for KL regularization that result in an entirely superior learning process to fine-tuning without it where plasticity to the new task is maintained while forgetting of general knowledge is greatly reduced. 

Another long-standing approach to eliminating forgetting in neural networks is experience replay \citep{Murre92,Lin92,Robins95}, which has been shown effective and computationally efficient in the context of LLMs as well \citep{abbes2025revisiting,li2025tic}. However, pure experience replay does not feel practical in the current age of open source LLMs as, in nearly all cases, open weight LLM models do not also publish copies of the data used for training. Part of the underlying reason for this is the use of proprietary data, and the internal value of data based on the work of paid annotators, which corporations are then less willing to share. Moreover, various licensing issues and privacy issues may arise from publishing the data used for pre-training models. As such, in this work we consider a practical alternative that we call \textit{approximate replay} where replay samples are drawn that mirror the next token prediction data seen during pre-training while leveraging an open web data corpus that is different than the actual corpus used for training. We find that despite the disconnect in data sources, approximate replay based on this open web data is still very effective at minimizing forgetting during fine-tuning without sacrificing plasticity to the new task. Indeed, the combined approach of \textit{approximate regularized replay} that utilizes both approximate replay and KL regularization provides an exceedingly simple yet computationally efficient solution for mitigating forgetting without diminishing the effectiveness of fine-tuning.

\section{Efficient and Stable Fine-Tuning for Large Language Models}

In general, the computational cost (in FLOPs) associated with fine-tuning a model $\theta$ on $N$ updates with a batch size of $B$ and context window size of $W$ can be expressed as $c_\text{FT} = 2NF_{\theta,B,W}$ where $F_{\theta,B,W}$ is the cost of forward propagation. This is because backward propagation over the full set of parameters $\theta$ has the same cost as the forward pass and both are needed to compute gradients.  

\subsection{LoRA}

Often full fine-tuning is not necessary though. Indeed, adapter models where some new parameters $\phi$ are learned to alter $\theta$ (and $|\phi|<<|\theta|$) have become quite popular as a method for fine-tuning LLMs. For example, in LoRA \citep{hu2022lora} a low-rank approximation of each weight matrix is learned. This can be very efficient in terms of parameters. In our experiments, as an example, $|\phi|$ is always less than 0.5\% the size of $|\theta|$. As $|\phi|<<|\theta|$, then the cost of backpropagation becomes insignificant relative to the cost of forward propagation and $c_\text{LoRA} \rightarrow NF_{\theta,B,W}$ in the limit of very few adapter parameters. In general, $NF_{\theta,B,W} \leq c_\text{LoRA} \leq 2NF_{\theta,B,W}$.

\textbf{Ability to Overwrite.} While for LoRA $|\phi|<<|\theta|$, it is important to note that LoRA models still have the ability to impact every parameter of the model $\theta$. Indeed, a LoRA model can be merged with the base model across all parameters by multiplying the low rank matrices \citep{hu2022lora}. As such, LoRA models are just as prone to overwriting knowledge in the model as full fine-tuning despite computational and parameter efficiency. 

\subsection{Stability and Plasticity During LLM Fine-Tuning}

Fine-tuning models on only a single task and expecting that the model performs well across a variety of tasks is unlikely to work due to the resulting biased optimization. This biased optimization is well characterized within the formalism of reinforcement learning through the conceptual framework of mixing times \citep{polynomial,riemerbalancing} and can be directly applied in supervised learning contexts as well \citep{crlsurvey}. We will now summarize some of these high-level insights to contextualize how the stability-plasticity dilemma \citep{StabilityPlasticity} arises. 

We can consider the current fine-tuning task as constituting a data distribution $d_\text{current}(x,y^*)$ over pairs of input contexts $x$ and associated optimal outputs $y^*$. Standard fine-tuning with LoRA then optimizes the objective $J^\text{current}(\theta,\phi) = \mathbb{E}_{x,y^* \sim d_\text{current}}[ \mathcal{L}_{\theta,\phi}^\text{SFT}(x,y^*)]$. However, the objective that we really care about evaluating our model on is the steady-state distribution which performs i.i.d. sampling over all future experiences $d_\text{future}(x,y^*)$ with associated objective $J^\text{eval}(\theta,\phi) = \mathbb{E}_{x,y^* \sim d_\text{future}}[ \mathcal{L}_{\theta,\phi}^\text{SFT}(x,y^*)]$. The problem is this distribution is generally unknowable. We can say it may bear some resemblance to $d_\text{current}(x,y^*)$ and it also may bear resemblance to an i.i.d. sampling over all past experiences $d_\text{past}(x,y^*)$. Moreover, there may be additional entirely novel experiences. So the stability-plasticity dilemma arises from uncertainty about the correct balance of current and past experiences to prepare the model as much as possible for the future. Concretely, plasticity measures progress on $J^\text{current}(\theta,\phi)$ and stability targets preservation or improvement of $J^\text{past}(\theta,\phi) = \mathbb{E}_{x,y^* \sim d_\text{past}}[ \mathcal{L}_{\theta,\phi}^\text{SFT}(x,y^*)]$. The reason why catastrophic forgetting occurs during fine-tuning is because of the distributional mismatch between $d^\text{current}(x,y^*)$ and $d^\text{past}(x,y^*)$ such that the more steps of consecutive optimization steps we take on $J^\text{current}(\theta,\phi)$, the less likely it is that this optimization also aligns with $J^\text{past}(\theta,\phi)$. Of course forgetting doesn't actually matter when it isn't relevant to $d_\text{future}(x,y^*)$, but when we optimize for multiple steps in a row on $d_\text{current}(x,y^*)$ we are implicitly conveying to the model that $d_\text{current}(x,y^*)$ is the future distribution even when this is only partially true. As such, we must consider ways to bias the optimization process towards learning the new task (i.e. promote plasticity) without disrupting the general capabilities of the model (i.e. while maintaining stability). 

\subsection{Methods to Promote Stability within LLMs During Fine-Tuning}

We consider two approaches for biasing optimization in favor of stability in this work: the KL divergence with respect to the base model and an approximation of replay from the pre-training phase using open data. We find that these two approaches are both very computationally efficient while providing significant leverage over balancing the stability-plasticity tradeoff. Moreover, because practitioners have control of the number of LoRA parameters and the replay rate, they can customize the compute overhead added to standard LoRA fine-tuning to meet use case requirements with more compute often leading to better results. 

\subsubsection{KL Regularization} \label{sec:method-kl}

KL regularization is a theoretically appealing and simple approach for allowing the initial model parameters to serve as a Bayesian prior on learning while making the model more robust in the face of spurious features. Given a given context $x$, we can express the base LLM's probability of producing output $y$ as $\pi_\theta(y|x)$. We can then express the output probabilities for the learned LoRA adapter on top of the base LLM as $\pi_{\theta+\phi}(y|x)$. If we want to promote stability in the model during fine-tuning, rather than optimizing $\phi$ directly for $\mathbb{E}_{x,y^* \sim d_\text{current}}[\mathcal{L}_{\theta,\phi}^\text{SFT}(x,y^*)]$, we can optimize for the KL regularized objective: $\mathbb{E}_{x,y^* \sim d_\text{current}}[\mathcal{L}_{\theta,\phi}^\text{SFT}(x,y^*) + \beta D_{\theta,\phi}^\text{KL}( \pi_{\theta+\phi}(\cdot|x) || \pi_{\theta}(\cdot|x))]$. Here $\beta$ is the KL regularization coefficient, which controls the degree of penalty when drifting from the output probabilities of the base model. Tuning $\beta$ thus provides directly leverage over the stability-plasticity tradeoff. High values of $\beta$ prevent forgetting while also preventing plasticity. Low values of $\beta$ allow for greater degrees of forgetting for old tasks and plasticity with respect to the new task. 

\textbf{Computational Overhead.} When applied to standard fine-tuning, KL divergence adds a $1.5\times$ computational overhead as it also requires an additional forward propagation through the initial model. However, there is significant synergy between computing KL divergence and using LoRA. Because the model being trained is only a LoRA adapter of the original model, it is then possible to perform both forward propagations with little computational overhead when $|\phi|<<|\theta|$. 

\textbf{Memory Overhead.} Once again there is synergy between the memory overhead of computing the KL divergence with respect to the base model and LoRA. During standard fine-tuning with KL regularization, it would be required to store two copies of the base model of size $|\theta|$ in memory. However, with LoRA it is only required that you store $|\theta|+|\phi|$, which adds no memory overhead over standard LoRA fine-tuning. 



\subsubsection{Approximate Experience Replay}

Another theoretically appealing and simple to implement stability prior is experience replay \citep{Murre92,Lin92,Robins95} in which past experiences are interleaved with current experiences during learning. If the environment follows a potentially unknown Markov chain, experience replay provides a nice theoretical solution. The experience replay buffer eventually converges to the steady-state distribution of the encountered contexts, which directly enables the model to combat optimization bias. However, practically speaking open source models are typically not released with the actual training data used. As a result, in this work we approximate replay by using an open source web corpus \url{https://huggingface.co/datasets/Skylion007/openwebtext}. The idea is that open source LLMs are trained on a large segment of web data using the next token prediction objective and that we can use that objective on a similar corpus to approximate the effect of a true experience replay implementation. We only leverage a very small segment of this corpus, so it would seem that random data should be representative and not present a tremendous mismatch with what was seen during training. 

\textbf{Computational Overhead.} In this setting, we can consider replay as equivalent to augmenting the fine-tuning dataset with more examples drawn from an open web corpus. Concretely, we define a replay rate $\rho$, which describes the amount of next token prediction replay examples of the given maximum context window $W$ for each example in the dataset. For example, $\rho=0$ corresponds to standard fine-tuning without replay, and $\rho=1$ corresponds to adding one replay example for each example in the fine-tuning datasets. As such, training with replay takes $(\rho+1)\times$ the amount of compute of standard fine-tuning and $(\rho+1)NF_{\theta,B,W} \leq c_\text{LoRA-Replay} \leq 2(\rho+1)NF_{\theta,B,W}$. 

\textbf{Memory Overhead.} Approximate replay requires $(\rho+1)\times$ more disk space to store the data, but does not necessarily require additional RAM on the CPU or GPU. 

\subsection{Theoretical Perspectives on Regularized Replay} \label{sec:theory}

Now that we have outlined the approach, we can expand on the theoretical perspective of what it achieves: 

\subsubsection{Why KL Regularization Prevents Forgetting}

\textbf{KL Regularization as a Bayesian Prior.}
From Bayesian perspective, learning can be treated as Maximum A Posteriori (MAP) estimation, from which we obtain the most likely model given both prior beliefs and the data from a new task. To see this, we treat the base-model $\pi_{\theta}$ as the prior distribution and take the likelihood as $p(\mathcal{D}|\phi)=\prod_{(x,y)\in \mathcal{D}}\pi_{\theta+\phi}(y|x)$. The posterior $p(\phi|\mathcal{D})$ is the updated belief (the fine-tuned model) after seeing the data. According to the Bayesian rule, $p(\phi|\mathcal{D})\propto p(\mathcal{D}|\phi)P(\phi)$. Taking the logarithm, we have $logP(\phi|\mathcal{D}) = \log p(\mathcal{D}|\phi) + \log p(\phi) + const$. In the context of KL regularization, $\log p(\mathcal{D}|\phi)$ is the standard cross-entropy loss ($\log\sum_{(x,y)\in \mathcal{D}}\pi_{\theta+\phi}(y|x)$) and the negative KL term $-\beta D_{KL}(\pi_{\theta+\phi}||\pi_{\theta})$ is equivalent to the log-prior. To obtain a valid Bayesian prior, $p(\phi)$ must be a proper probability distribution, if we exponentiate the log-prior, we obtain $p(\phi) = exp(-\beta D_{KL}(\pi_{\theta+\phi}||\pi_{\theta})/Z$, which is known as a Boltzmann distribution (or Gibbs distribution) over the space of models with $\beta^{-1}$ being the temperature and $Z$ the partition function. This prior explicitly encodes the belief that the most likely model is the base model, and the probability of any other model decays exponentially as its output distribution divergences from the base model. To further understand how a prior on the outputs relates to the weights $\theta+\phi$, we can use a second-order Taylor expansion to approximate the KL divergence. For a small change in weights $\Delta\theta = \phi = \theta' - \theta$, the KL divergence is approximately: $D_{KL}(\pi_{\theta+\phi}||\pi_{\theta})\approx(\theta'-\theta)^TF(\theta)(\theta'-\theta)=\phi^TF(\theta)\phi$, where $F(\theta)$ is the Fisher information matrix. By substituting this approximation back into the log-prior, we have $\log p(\phi)\approx -\frac{\beta}{2}\phi^TF(\theta)\phi$, which is exactly the log-density of a multivariate Gaussian distribution: $p(\phi)=\mathcal{N}(0, \frac{1}{\beta}F(\theta)^{-1})$.

\subsubsection{Why KL Regularization Can Improve Plasticity}  \label{sec:theory-bottleneck}

\textbf{KL Regularization and Robustness to Spurious Features.} Supervised fine-tuning with KL regularization and LoRA adapters can also be theoretically interpreted through the lens of the information bottleneck (IB) objective \citep{deng2025enhancing,Zhuang2025Dynamic}: $I(Y;\pi_{\theta+\phi})-{\beta}I(X; \pi_{\theta+\phi})$, where the log-likelihood term maximizes $I(Y;\pi_{\theta+\phi})$, the mutual information between output $y$ and $\pi_{\theta+\phi}$ (model fitting objective) and the KL term is the upper bound on $I(X; \pi_{\theta+\phi})$, the mutual information between context $x$ and $\pi_{\theta+\phi}$ (model compression objective). The IB framework~\cite{Tishby2015IB} aims to minimize $I(X; \pi_{\theta+\phi})$ the information retained about the input $x$ to promote generalization. However, directly computing $I(X; \pi_{\theta+\phi})$ is computationally intractable for neural networks. To address this issue, one typically applies the variational upper bound of $I(X; \pi_{\theta+\phi})$ derived as $I(X; \pi_{\theta+\phi})\leq\mathbb{E}[D_{KL}(\pi_{\theta+\phi}||\pi_{\phi})]$ as a surrogate~\cite{alemi2016deep}. It can be shown that, by minimizing the KL divergence to a reference model, one can effectively minimize the mutual information between the input prompt and the model's internal state~\cite{lei2025revisiting}. This motivates learning to ignore spurious features in the input prompt (such as specific phrasing or noise) and only keep the essential features needed to generate the correct answer.

\subsubsection{Why Replay Stabilizes Learning}

\textbf{Replay and Steady-State Optimization.} In supervised learning, it is typically assumed that an agent's behavior does not impact future experiences. In this case, it is possible to model the agent's experiences as some unknown Markov chain $P(x_{t+1},y^*_{t+1}|x_{t},y^*_{t})$ where $x_{t}$ is the current context and $y^*_{t}$ is the associated optimal output. While our supervised fine-tuning dataset only represents $BN$ steps from this chain, what the agent really cares to optimize over is all steps that it will encounter in it's lifetime. In fact, it is the disconnect between these two distributions that is the underlying cause of catastrophic forgetting. If the lifetime is sufficiently large (i.e. greater than the mixing time of the chain) then this converges to a steady-state distribution $d_\text{future}(x,y^*)$ over which we want to minimize $\mathcal{L}_{\theta,\phi}^\text{SFT}(x,y^*)$. Replay provides at least an asymptotic solution to this problem without attempting to model the Markov chain directly. This is because as a replay buffer fills, the sampling distribution from the buffer will asymptotically converge to $d_\text{future}(x,y^*)$. 

\section{Experimental Setup}

\textbf{Training Datasets.} For the training tasks, we consider a set of 5 tasks inspired by the prior work on catastrophic forgetting during supervised fine-tuning of \citet{luo2025empirical} in which the authors had selected from a subset of the instruction following tasks considered by \citet{scialom2022fine}:   

\begin{enumerate}[leftmargin=*]
    \item \textbf{Text Simplification (Simp):} This task requires the LLM to paraphrase the provided text with a simple shorter piece of text \citep{jiang2020neural,alva2020asset}. Concretely, the model is instructed to "Reformulate this text with simpler words: " where the normal article text and simplified article text (as a target for supervision) are provided by part 1 (for training) and part 2 (for testing) of the dataset \url{https://huggingface.co/datasets/chaojiang06/wiki_auto}. 
    \item \textbf{Empathetic Dialogue Generation (Emdg):} This task requires the LLM to generate a response to a conversational context under a given emotional situation and was previously considered by \citet{rashkin2019towards}. Concretely, the model is given an instruction of the form "The associated emotion is \{emotion\} and the input prompt is \{prompt\}. Now what would be your response, given the following dialogue context:===\{text\}". The training and testing data is pulled from the splits provided at the repository \url{https://huggingface.co/datasets/facebook/empathetic_dialogues}.
    \item \textbf{Inquisitive Question Generation (InqQG):} This task requires the LLM to generate a simple question that could be associated with a long-form answer and was previously considered by \citet{fan2019eli5}. Concretely, the model is given an instruction of the form "\{text\}===Given the above text, write the possible curious question it answers:". The training and testing data is drawn from the splits provided by the repository \url{https://huggingface.co/datasets/Pavithree/eli5}. 
    \item \textbf{Explanation Generation (Exp):} This task requires the LLM to generate an explanation about why two sentences are different and was previously considered by \citet{camburu2018snli}. Concretely, the model is instructed to "Explain why the two following sentences are unrelated: Sentence 1: \{first-sentence\}; Sentence 2: \{second-sentence\}". The data is sampled from both training splits and the testing split of the repository \url{https://raw.githubusercontent.com/OanaMariaCamburu/e-SNLI/master/dataset/}.
    
    \item \textbf{Headline Generation with Constraint (HGen):} This task requires the LLM to generate headlines for articles and was previously considered by \citet{scialom2022fine}. However, the data used by \citet{scialom2022fine} and \citet{luo2025empirical} requires an LDC license, so we opt for the dataset of \citet{leeb2024diverse} to allow for greater general purpose reproducibility. Concretely, the model is instructed to "Make a title for this article: \{article\}". The training and testing data consist of random subsets of the english titles and articles from the repository \url{https://huggingface.co/datasets/felixludos/babel-briefings} where data is filtered such that the article is at least $3\times$ longer than the title and the title is at least 3 words. 
\end{enumerate}

\textbf{Training Procedure.} Our training procedure was implemented by extending the Transformers Trainer class and deployed across a cluster of H100 GPUs. We found that the AdamW optimizer with a constant learning rate achieved the same performance as a cosine schedule with warm-up and chose a constant learning rate without warm-up for simplicity and to stay consistent with \citet{luo2025empirical}. We followed \citet{luo2025empirical} and set the context length for these tasks to 512. Based on our preliminary runs with the 3B model, we set the LoRA rank $r=32$ and $\alpha=64$ such that $\alpha/r=2$, the learning rate to $1e-4$, the LoRA dropout rate to 0.05, and the batch size to $8$ for all experiments. For each task, we sample $1,000$ random examples from the training set and $1,000$ random examples from the testing set. Our experiments ran on from $1$ to $4$ H100 GPUs at a time, depending on the model size, in order to make sure we satisfied GPU memory requirements. All reported results are an average of 7 random seeds for each task and hyperparameter combination. 

\textbf{Model Sizes.} We consider a variety of model sizes within the Qwen 2.5 Instruct \citep{qwen2025qwen25technicalreport} family of models. Specifically, we ran all experiments across the 1.5B, 3B, 7B, and 14B instruction-tuned models. We build LoRA adapters for the key, value, and output matrices. This corresponds to trainable parameters that are $0.46\%$ the size of the base model for the 1.5B model, $0.39\%$ the size of the base model for the 3B model, $0.22\%$ the size of the base model for the 7B model,  and $0.28\%$ the size of the base model for the 14B model. 

\textbf{Evaluating Plasticity.} In order to assess the plasticity or adaptation performance of each model to the task it is being trained on we evaluate on the held out test set data for each task. While previous papers considered different metrics catered to each task \citep{scialom2022fine,luo2025empirical}, we found this made it difficult to assess average across task performance fairly. As a result, we opted for the simple solution of always evaluating performance based on the BERTScore \citep{zhang2019bertscore} F1 metric between the generate response and gold label on the testing data. Each model achieved around an $81$ average performance across the 5 tasks by this metric prior to training. We denote the amount of plasticity as $\uparrow P$ in our experiments, which is the average score after training subtracted by the average score before training. We use $\uparrow$ to indicate that higher scores are better with larger values indicating improvement of the ability to generalize to held out examples from the fine-tuning task and negative values indicated that training actually had a counter productive effect. 

\textbf{Evaluating Forgetting.} We follow the procedure established by \citet{shenfeld2025rl} for a general purpose evaluation of knowledge in the LLMs across a variety of capabilities to assess catastrophic forgetting. Leveraging the lm-evaluation-harness (\url{https://github.com/EleutherAI/lm-evaluation-harness}), we evaluate each model before and after training on the average of six datasets: HellaSwag, HumanEval, IFEval, MMLU, TruthfulQA, and WinoGrande. The average scores before training for each model were $48.13$ for the 1.5B model, $58.37$ for the 3B model, $66.22$ for the 7B model, and $68.31$ for the 14B model. We denote the amount of forgetting as $\downarrow F$ in our experiments, which is the average score after training subtracted from the average score before training. We use $\downarrow$ to indicate that lower scores are better i.e. negative forgetting is positive transfer and indicates improvement of the general capabilities of the model during training. 

\section{Empirical Results}

We provide the main results of our comprehensive experiments across base model sizes $|\theta|$, replay rates $\rho$, and KL coefficients $\beta$ in Tables \ref{absolute-table} and \ref{relative-table}. The first row of Table \ref{absolute-table} reveals the very significant catastrophic forgetting involved in standard LoRA fine-tuning. Interestingly, forgetting seems to get even worse with increased model size in this regime. One potential explanation would be that the larger models have better initial performance and more to lose, but our results in Table \ref{relative-table} demonstrate that the relative loss of performance also grows. 

\textbf{Contextualizing How Catastrophic Forgetting Is.} One way to understand the effect of forgetting is in terms of how final general performance compares to other smaller models. Indeed, the 3B model after training performs worse than the 1.5B model, the 7B model after training is comparable to the performance of the 1.5B model, and the 14B model after training is even worse than the performance of the 1.5B model. As such, the effect of forgetting is equivalent to downgrading the model significantly in terms of general capabilities such that it is comparable to a significant loss of parameters.  

\textbf{The Effect of Approximate Replay.} Our experiments reveal that approximate replay provides a significant deterrent to forgetting while retaining the ability to achieve the plasticity of standard fine-tuning. Indeed, on average approximate replay alone provides about a $3\times$ reduction in the amount of forgetting without sacrificing plasticity. It does appear, however, that the effectiveness of replay degrades with more investment, particularly in terms of the computational overhead. We achieve the best performance with a replay rate of 3X, but 1X provides the most economical solution if there are constraints on the compute.  

\textbf{The Effect of KL Divergence.} Our experiments also demonstrate the ability to manipulate the stability-plasticity tradeoff by setting an appropriate KL coefficient. $\beta=0.1$ seems to provide a very substantial deterrent for changing the model parameters. This results in a virtual elimination of forgetting, but also all but eliminates the plasticity of the model. $\beta=0.01$ seems to provide a better tradeoff. Forgetting is still virtually eliminated, and while plasticity is worse than standard fine-tuning, it is not that much worse. $\beta=0.001$ allows for much more flexibility in the model and achieves plasticity that even slightly surpasses standard fine-tuning. This improved generalization to the new task makes sense given our remarks in Section \ref{sec:theory-bottleneck}. However, $\beta=0.001$ also allows for a significant degree of forgetting. That said, $\beta=0.001$ represents an entirely improved solution over standard fine-tuning with minimal computational and memory overhead as it still improves on forgetting substantially over standard fine-tuning. 

\textbf{Combining Replay and KL Divergence.} The best results come from combining approximate replay with KL divergence regularization. For example, replay is able to improve even further on $\beta=0.001$ by retaining plasticity while cutting down even more on the extent of forgetting. Overall, the best combination depends on the perceived tradeoff between stability and plasticity. Approximate replay with $\beta=0.01$ provides virtually no forgetting while experiencing only a mild loss in term of plasticity in comparison to standard fine-tuning. On the other hand, approximate replay with $\beta=0.001$ provides the same plasticity as standard fine-tuning with an over $7\times$ average reduction in the amount of forgetting experienced. 

\begin{table}[t]
\caption{Average Absolute Performance Change After Training}
\label{absolute-table}
\begin{center}
\begin{NiceTabular}{cc|cccc|c}
\multicolumn{1}{c}{\bf Replay} & \multicolumn{1}{c}{\bf KL} &\multicolumn{1}{c}{\bf 1.5B} &\multicolumn{1}{c}{\bf 3B} &\multicolumn{1}{c}{\bf 7B} &\multicolumn{1}{c}{\bf 14B} & \multicolumn{1}{c}{\bf Average} \\ 
\multicolumn{1}{c}{\bf Rate $\rho$} & \multicolumn{1}{c}{\bf Coeff $\beta$} & $\downarrow$ F  / $\uparrow$ P & $\downarrow$ F  / $\uparrow$ P  & $\downarrow$ F  / $\uparrow$ P & $\downarrow$ F  / $\uparrow$ P & $\downarrow$ F  / $\uparrow$ P  \\ \hline 
0X & 0.0 & 9.3 / 2.3  & 14.4 / 2.1  & 17.1 / 2.6  & 20.9 / 2.9  & 15.4 / 2.5  \\
0X & 0.1 & 0.5 / 0.3 & -0.1 / 0.4  & -0.2 / 1.1 & -0.3 / 1.0 & 0.0 / 0.7 \\
0X & 0.01 & 1.8 / 1.5 & 0.1 / 1.8 & 1.0 / 2.2 & 2.0 / 1.7 & 1.2 / 1.8 \\
0X & 0.001 & 5.1 / 2.5 & 6.7 / 2.6 & 9.7 / 2.7& 11.9 / 3.0 & 8.4 / 2.7  \\ \hline
1X & 0.0 & 2.2 / 2.4 & 4.1 / 1.8 & 6.8 / 2.6 & 8.5 / 2.7 & 5.4 / 2.4 \\
1X & 0.1 & 0.2 / 0.1 & -0.3 / 0.4 & -0.3 / 0.9 & -0.4 / 0.3 & -0.2 / 0.4  \\
1X & 0.01 & 0.4 / 1.2 & -0.7 / 1.5 & -0.1 / 2.0 & 0.0 / 1.3 & -0.1 / 1.5  \\
1X & 0.001 & 1.3 / 2.4 & 0.7 / 2.5 & 3.7 / 2.7 & 5.3 / 2.9 & 2.8 / 2.6 \\ \hline
3X & 0.0 & 1.8 / 2.5 & 3.9 / 2.1  & 5.2 / 2.6 & 7.4 / 2.6 & 4.6 / 2.7\\
3X & 0.1 & 0.2 / 0.1 & -0.5 / 0.4 & -0.1 / 0.8 & 0.1 / 0.2 & -0.1 / 0.4  \\
3X & 0.01 & 0.4 / 1.1 & -0.5 / 1.6 & 0.0 / 2.0 & -0.2 / 2.1 & -0.1 / 1.7  \\
3X & 0.001 & 0.9 / 2.6 & 0.6 / 2.6 & 2.6 / 2.8 & 3.8 / 2.9 & 2.0 / 2.7 \\ \hline
7X & 0.0 & 1.5 / 2.3 & 4.2 / 2.2 & 6.3 / 2.6 & 7.5 / 2.8 & 4.9 / 2.5  \\
7X & 0.1 & 0.0 / 0.1 & -0.4 / 0.4 & 1.3 / 0.7 & -0.2 / 0.3 & 0.2 / 0.4   \\
7X & 0.01 & 0.3 / 1.2 & -0.2 / 1.7 & 0.1 / 1.9 & 0.1 / 1.7 & 0.1 / 1.6 \\
7X & 0.001 & 1.0 / 2.5 & 0.7 / 2.6 & 2.8 / 2.7 & 3.3 / 3.0 & 2.0 / 2.7 \\ \hline
\end{NiceTabular}
\end{center}
\end{table}

\begin{table}[t]
\caption{Average Relative Performance Change After Training}
\label{relative-table}
\begin{center}
\begin{NiceTabular}{cc|cccc|c}
\multicolumn{1}{c}{\bf Replay} & \multicolumn{1}{c}{\bf KL} &\multicolumn{1}{c}{\bf 1.5B} &\multicolumn{1}{c}{\bf 3B} &\multicolumn{1}{c}{\bf 7B} &\multicolumn{1}{c}{\bf 14B} & \multicolumn{1}{c}{\bf Average} \\ 
\multicolumn{1}{c}{\bf Rate $\rho$} & \multicolumn{1}{c}{\bf Coeff $\beta$} & $\downarrow$ F  / $\uparrow$ P & $\downarrow$ F  / $\uparrow$ P  & $\downarrow$ F  / $\uparrow$ P & $\downarrow$ F  / $\uparrow$ P & $\downarrow$ F  / $\uparrow$ P  \\ \hline 
0X & 0.0 & 19.2\% / 2.9\%  & 24.7\% / 2.6\%  & 25.9\% / 3.3\%  & 30.6\% / 3.6\%  & 25.1\% / 3.1\%    \\
0X & 0.1 & 0.9\% / 0.4\% & -0.2\% / 0.5\% & -0.3\% / 1.4\% & -0.4\% / 1.2\% & 0.0\% / 0.9\%  \\
0X & 0.01 & 3.8\% / 1.8\% & 0.2\% / 2.3\% & 1.5\% / 2.8\% & 2.9\% / 2.2\% & 2.1\% / 2.3\% \\
0X & 0.001 & 10.5\% / 3.1\% & 11.4\% / 3.2\% & 14.7\% / 3.4\% & 17.4\% / 3.7\% & 13.5\% / 3.4\%  \\ \hline
1X & 0.0 & 4.6\% / 3.0\% & 7.0\% / 2.3\% & 10.3\% / 3.2\% & 12.5\% / 3.4\% & 8.6\% / 3.0\%  \\
1X & 0.1 & 0.5\% / 0.2\% & -0.6\% / 0.4\% & -0.4\% / 1.1\% & -0.6\% / 0.4\% & -0.3\% / 0.5\% \\
1X & 0.01 & 0.9\% / 1.5\% & -1.2\% / 1.9\% & -0.1\% / 2.5\% & 0.0\% / 1.6\% & -0.1\% / 1.9\% \\
1X & 0.001 & 2.7\% / 2.9\% &  1.2\% / 3.1\% & 5.7\% / 3.4\% &  7.7\% / 3.6\% & 4.3\% / 3.3\% \\ \hline
3X & 0.0 & 3.8\% / 3.0\%  & 6.6\% / 2.6\% & 7.8\% / 3.3\% & 10.3\% / 3.3\% & 7.1\% / 3.1\%   \\
3X & 0.1 & 0.3\% / 0.1\% & -0.9\% / 0.5\% & -0.3\% / 1.0\% & 0.2\% / 0.2\% & -0.2\% / 0.5\%  \\
3X & 0.01 & 0.8\% / 1.4\% & -0.9\% / 2.0\% & 0.0\% / 2.5\% & 0.2\% / 2.5\% & 0.0\% / 2.1\% \\
3X & 0.001 & 1.8\% / 3.2\% & 1.0\% / 3.2\% & 4.0\% / 3.4\% & 5.5\% / 3.6\% & 3.1\% / 3.4\% \\ \hline
7X & 0.0 & 3.2\% / 2.8\% & 7.3\% / 2.7\% & 9.6\% / 3.2\% & 11.0\% / 3.5\% & 7.8\% / 3.1\% \\
7X & 0.1 & 0.0\% / 0.2\% & -0.7\% / 0.5\% &  1.9\% / 0.8\% & -0.2\% / 0.4\% & 0.3\% / 0.5\% \\
7X & 0.01 & 0.7\% / 1.4\% & -0.4\% / 2.1\% & 0.1\% / 2.3\% &  0.1\% / 2.1\% & 0.1\% / 2.0\% \\
7X & 0.001 & 2.1\% / 3.0\% & 1.1\% / 3.2\% & 4.3\% / 3.3\% & 4.8\% / 3.7\% & 3.1\% / 3.3\% \\ \hline
\end{NiceTabular}
\end{center}
\end{table}


\section{Related Work}

Our work is related to a variety of directions of study in the continual learning literature. 

\textbf{KL Regularization in RL.} As mentioned earlier, KL regularization of the form used in our paper has become commonplace when performing RL with LLMs. In particular, it is generally applied in concert with PPO \citep{Ouyang2022InstructGPT,Liu2025RethinkingKLRegularization}. We argue in this paper that it should also be widely used during supervised fine-tuning. 

\textbf{Connections to Distillation.} Our use of KL regularization during learning also bears similarities to prior work leveraging distillation to aid with continual learning both with \citep{buzzega2020dark} and without \citep{riemer2017representation,LwF} replay buffers. KL regularization can be seen as a particularly simple form of distillation that focuses only on output probabilities rather than differences at the hidden layers. 

\textbf{Replay Buffer Types.} The approximate replay buffer we consider in this work is theoretically related to reservoir sampling based buffers \citep{RS,MER} in that it draws a random subset over the data of a prespecified size. A recency based sampling \citep{Mnih2013} wouldn't make sense in our setting as it would be likely to sample correlated data that would make learning less robust. We also experimented with generative replay approaches \citep{shin2017continual,GenerativeDist,riemer2019scalable,bashivan2019continual} based on examples generated by the LLM itself, but found it difficult to get sufficient diversity in the generated experiences to make a meaningful difference in stabilizing learning. 

\textbf{Sparse Modular Models.} Architectures that exploit modular structure such as sparse mixtures of experts (MoE) \citep{Hinton91,Miikkulainen1993,Jordan94,davis2013low} have established benefits in the field of continual learning \citep{riemer2015deep,CrossStitch,Pathnet} and have become commonly used for training LLMs due to their computational advantages \citep{largeneuralnets}. As discussed by \citet{rosenbaum2019routing}, modular architectures with dynamic composition \citep{rosenbaum2018routing,cases2019recursive,chang2018automatically,rosenbaum2019dispatched,thomas,kostas20a,zini2020coagent} have the ability to effect the dynamics of transfer and forgetting by allowing the model to orthogonalize weight updates by routing experiences to different modules. That said, routers are not necessarily trained to make routing decisions with gradient interference in mind and may fail to live up to this promise in practice. \citet{therien2025continual} recently explored the influence of the MoE router in effecting the dynamics of continual pre-training. We avoid the use of MoE models in our experiments to avoid this potential conflating factor in the results, but believe this is a promising direction for improving fine-tuning of LLMs while preventing forgetting.

\textbf{Model Merging.} Another interesting approach for preventing forgetting in LLMs is model merging \citep{wortsman2022model,ilharco2022editing,yadav2024ties,akiba2024evolutionary} with recent work exploring model merging in the context of continual learning \citep{phan2025toward}. LLMs fine-tuned using LoRA could be merged with the base model or other task specific models to improve retention of general knowledge \citep{thakkar2024deep,thakkar-etal-2025-combining}. This approach is complementary to the direction considered in our work. 

\textbf{Capacity Regularization.} The KL loss and approximate replay both serve to regularize the learning objective during fine-tuning. Another interesting form of regularization is to limit the capacity of the model to acquire potentially spurious knowledge \citep{malloy2020deep,malloy2020consolidation,malloy2021capacity,malloy2021rl,malloy2023learning}. As we show in Section \ref{sec:theory-bottleneck}, the IB theory suggests that we should get this benefit for free when using KL regularization as we do in this work. 

\textbf{The Impact of Model Size on Forgetting.} An interesting aspect of our results is that we find catastrophic forgetting in fine-tuning to be even worse for bigger models than it is for smaller models. This directly contradicts findings in the work of \citet{ramasesh2022effect} suggesting that larger models experience less forgetting than small models. Larger models experiencing less forgetting seems to be generally true with large datasets such as during continual pre-training \citep{abbes2025revisiting}. However, the numbers in \citet{luo2025empirical} interestingly also suggest that the effect may be the opposite during LLM fine-tuning. In our work we consider a wider array of model sizes and see this result of more forgetting in larger models more consistently and with a greater effect size. This result serves as a cautionary tale to practitioners who may be comforted by the argument of \citet{ramasesh2022effect} and not worried about forgetting during fine-tuning because they use large models.

\section{Discussion}

In this paper, we have proposed a very simple yet efficient and effective strategy for stabilizing learning during LLM fine-tuning. The potential applications of this approach are vast and we refer readers to Appendix \ref{app:applications} for an in-depth discussion. Our proposal of approximate regularized replay combines two straightforward approaches in the continual learning literature in KL regularization and experience replay to largely eliminate forgetting during LLM fine-tuning without sacrificing the ability to adapt to the new fine-tuning task. Our approach prioritizes efficiency in leveraging parameter efficient tuning based on LoRA with a customizable degree of additional computational overhead that can be tuned to meet use case requirements. Moreover, our approach prioritizes practicality by leveraging an open source dataset that is used as a proxy for replay in lieu of direct access to the pre-training data used for the model. Our work takes a step in the direction of democratizing the ability to fine-tune LLMs towards specific business needs, which we believe may be a crucial bottleneck in achieving higher rates of success integrating generative AI across business applications. 

\subsubsection*{Acknowledgments}
We thank the IBM Cognitive Compute Cluster and IBM Blue Vela Cluster for providing computational resources for our experiments. MR would also like to acknowledge support from the Canada Excellence Research Chairs (CERC) Program.

\bibliography{main}
\bibliographystyle{tmlr}

\appendix

\section{Discussion of Promising Applications} \label{app:applications}

Here we provide an in-depth discussion of potential applications enabled by the simple and efficient continual fine-tuning approach we propose in this paper. 

\textbf{Instruction-Tuning.} In the experiments of our paper, we considered the use case of instruction-tuning language models based on supervised fine-tuning. This has long been a preeminent approach for improving and customizing LLMs \citep{wei2021finetuned,ouyang2022training,chung2022scaling,aribandi2022extensible} see \citep{zhang2023instruction,han2025alignment} for related surveys. It is particularly common in a business context when practitioners attempt to customize models to perform better at their business's specific use case with some generally limited data for supervision. Indeed, popular open source projects have picked up on this trend and have developed synthetic data generation approaches to help practitioners build more robust datasets for this fine-tuning \citep{sudalairaj2024lab}. Instruction-tuning is a very general purpose use case that can be applied in customizing AI models across industries spanning use cases in finance \citep{zhang2023instructfingpt,xie2023pixiu,tanabe2024jafin,fatemi2024comparative}, marketing \citep{wu2025grounded}, the social sciences \citep{dey2024socialite,wang2024css_instruction_tuning,khabiri2015domain,heath2015alexandria,riemer2017distributed}, and healthcare \citep{zhang2023alpacare,tran2023bioinstruct,gupta2024llamacare,das2018pepcvae} just to name a few.

\textbf{Real-time Learning.} Providing efficient continual learning is very important for real-time applications as efficient and stable gradient calculations cut down on the delay associated with updating the model \citep{anokhinhandling} and provides better models for asynchronous execution \citep{riemer2024realtime,riemer2025enabling}. 

\textbf{Theory of Mind.} It is also an important stepping stone in providing better theory of mind in LLMs \citep{ashktorab2025bridging}. Recent work has found that LLMs have a limited ability to have theory of mind impact their behavior based on in-context learning alone  \citep{riemer2024can,riemerposition}. Moreover, leveraging parameter efficient LoRA adapters helps mitigate scalability concerns involved in developing separate theory of mind models for each user \citep{memarian2022summarizing,touzel2024scalable}. 

\textbf{Customized Contextual Alignment of LLMs.} This parameter efficient approach also provides a potential solution to the contextual alignment problem \citep{padhi2024comvas,dognin2024contextual,dognin2025contextual} although it is important for practitioner to still consider the proper scope of alignment \citep{varshney2025scopes} as it is not always desirable to do exactly what an individual user wants if it results in potential externalities. 

\textbf{Multi-Agent Interaction.} Multi-agent environments are inherently non-stationary as other agents learn, which leads to the existence of \textit{active equilibria} \citep{foerster2017learning,kim2021policy,kim2022influencing,kim2022game} available when continual learning that are superior to any fixed policy Nash equilibria. In this setting, stable fine-tuning can even allow agents to learn from the actions and goals advised by other agents during interaction \citep{torrey2013teaching,taylor2014reinforcement,fachantidis2017learning,da2017simultaneously,omidshafiei2019learning,kim2019learning,kim2020heterogeneous}. 

\textbf{Learning Hierarchical Policies.} Supervised fine-tuning of adapters derived from the same base model could also be useful for building sub-policies within a modular agentic architecture as envisioned by \citet{miehling2025agentic}. The options framework \citep{sutton1999between} in RL then provides a principled framework for learning to coordinate the use of these sub-policies \citep{bacon2017option,riemer2018learning2,riemer2020role}. Alternatively, the LLM experts can learn to coordinate by conversing in natural language \citep{sun2023corex,liang2024encouraging,wang2024rethinking,eo2025debate}. While it is possible to learn everything in a single sufficiently complex generalist agent, there is a benefit in learning sub-policies in order to exploit the different state abstractions that they learn for improved generalization based on limited data \citep{abdulhai2022context}. Indeed, this may be a necessary ingredient in achieving compositional generalization across concepts \citep{klinger2020study}. Moreover, this approach has also been shown effective when aggregating across data sources for time-series prediction \citep{riemer2016correcting}. 

\end{document}